\documentclass[11pt,a4paper]{article}
\usepackage[hyperref]{acl2021}
\usepackage{times}
\usepackage{latexsym}

\usepackage{amsmath,amsfonts}
\usepackage[inline]{enumitem}

\usepackage{microtype}

\aclfinalcopy 
\def\aclpaperid{3598} 

\usepackage{graphicx}
\usepackage{color}

\title{Task-adaptive Pre-training of Language Models with Word Embedding Regularization}

\author{Kosuke Nishida,
		Kyosuke Nishida,
		Sen Yoshida\\
		\rm NTT Media Intelligence Laboratories, NTT Corporation\\
		\tt kosuke.nishida.ap@hco.ntt.co.jp}
\date{}

\begin{document}
\maketitle
\begin{abstract}
Pre-trained language models (PTLMs) acquire domain-independent linguistic knowledge through pre-training with massive textual resources. Additional pre-training is effective in adapting PTLMs to domains that are not well covered by the pre-training corpora. Here, we focus on the static word embeddings of PTLMs for domain adaptation to teach PTLMs domain-specific meanings of words. We propose a novel fine-tuning process: task-adaptive pre-training with word embedding regularization (TAPTER). TAPTER runs additional pre-training by making the static word embeddings of a PTLM close to the word embeddings obtained in the target domain with fastText.  TAPTER requires no additional corpus except for the training data of the downstream task. We confirmed that TAPTER improves the performance of the standard fine-tuning and the task-adaptive pre-training on BioASQ (question answering in the biomedical domain) and on SQuAD (the Wikipedia domain) when their pre-training corpora were not dominated by in-domain data.
\end{abstract}

\section{Introduction}
Pre-trained language models (PTLMs) trained with massive textual and computational resources have achieved high performance in natural language processing tasks \cite{bert}.
Additional pre-training often is used to tackle domain discrepancies between the downstream task and the pre-training corpora.
Additional pre-training with a large corpus in the domain of the downstream task, such as BioBERT~\cite{biobert}, improves the performance on the task \cite{clinicalbert, scibert, legalbert}.
However, this approach requires large corpora in the target domain and entails a high computational cost. 

\citet{tapt} proposed task-adaptive pre-training (TAPT), which is additional pre-training using only the training data of the downstream task. 
TAPT can be regarded as a new fine-tuning process in which 
the standard fine-tuning is preceded by low-cost additional pre-training.

In this study, we focus on the static word embeddings of PTLMs (i.e., non-contextualized 0-th layer representations)  for domain adaptation.
Our method is designed to teach PTLMs the domain-specific meanings of the words as static word embeddings.
We are motivated by the observation that the middle BERT layers capture the syntactic information \cite{middle1, middle2, middle3}. 
We consider that we can adapt the models without harming the domain-independent linguistic knowledge contained in higher layers by learning the static word embeddings directly.

We propose a novel fine-tuning process called task-adaptive pre-training with word embedding regularization (TAPTER).
First, TAPTER obtains word embeddings in the target domain by adapting a pre-trained fastText model \cite{fasttext1} to the target domain using the training data of the downstream task.
Next, TAPTER runs the task-adaptive 
pre-training by making the static word embeddings of the PTLM close to the word embeddings obtained with the fastText model. Finally, TAPTER runs the standard fine-tuning process.

We found that TAPTER achieves higher scores than the standard fine-tuning and TAPT on question answering tasks in the biomedical domain, BioASQ \cite{bioasq}, and Wikipedia domain, SQuAD1.1 \cite{squad}.
Our key findings are:
\begin{enumerate*}[label=(\roman*)]
\item Word embedding regularization in task-adaptive pre-training enhances domain adaptation when the initial pre-training corpora do not contain a high proportion of 
in-domain data.
\item The word embeddings of fastText, which uses a shallow neural network, can be adapted to the target domains more easily than the static word embeddings of PTLMs. 
\end{enumerate*}

\section{Preliminaries}
\subsection{Pre-trained Language Models}
We focus on the static word embeddings of PTLMs.
Let $V_{\mathrm{LM}}$ be the vocabulary.
We input a token sequence $X \in V_{\mathrm{LM}}^l$ to the model, where $l$ is the length of the sequence.
The embedding layer of the model has a word embedding matrix $E \in \mathbb{R}^{V_{\mathrm{LM}}\times d_{\mathrm{LM}}}$ as trainable parameters, where $d_{\mathrm{LM}}$ is the embedding dimension.
The word embedding of the $i$-th token is $E_{x_i}$.

The vocabulary of PTLMs consists of subword units; for example, 30K WordPiece tokens ~\cite{wordpiece} are used in BERT~\cite{bert}
and 50K byte-level BPE tokens~\cite{bpe} are used in RoBERTa~\cite{roberta}.


\subsection{fastText}
fastText is a word embedding method using subword information~ \cite{fasttext1}.
The skipgram model~\cite{word2vec} of fastText learns word embeddings by predicting the surrounding words $x_j~(j \in C_i)$ from a word $x_i$, where $C_i$ is the set of the indices within a given window size.
Specifically at position $i$, we use the surrounding words as positive examples and randomly sample negative words $\mathcal{N}_i$ from the vocabulary $V_{\mathrm{FT}}$.
The loss function is
\begingroup\makeatletter\def\f@size{9}\check@mathfonts
\[
\sum_i \left\{ 
\sum_{j \in C_i} \log (1+e^{-s(x_i, x_j)})
+\sum_{x \in \mathcal{N}_i} \log (1+e^{s(x_i, x)})
\right\}.
\]
\endgroup
That is, the model learns to score higher for positive examples and lower for negative examples.

fastText uses subword information to model the score function $s$.
Let $S_v$ be the set of substrings of the word $v \in V_{\mathrm{FT}}$.
The score of the input word $x_i$ and the output word $x_j$ is
\[
s(x_i, x_j) =\sum_{w \in S_{x_i}} W_{\mathrm{in},w}^\top W_{\mathrm{out},x_j}.
\]
Here, $W_{\mathrm{in}} \in \mathbb{R}^{N\times d_{\mathrm{FT}}}$ consists of the word embeddings of the input layer
and $W_{\mathrm{out}} \in \mathbb{R}^{V_{\mathrm{FT}}\times d_{\mathrm{FT}}}$ consists of the word embeddings of the output layer.
$d_{\mathrm{FT}}$ is the embedding dimension, and $N$ is an arbitrary large number that determines the actual vocabulary size of the subwords.
In the implementation of fastText, the model does not restrict the vocabulary size by hashing a subword $w$ into an index less than $N$. The model has limits on the minimum and maximum lengths of subwords.

At inference time, the embedding of a word $w$ is $\sum_{w \in S_{v}} W_{\mathrm{in},w}$.
\citet{fasttext1} reported that fastText learns word similarity with less training data than other methods do by utilizing the subword information.

\subsection{Related Work}
Static word embeddings in PTLMs have attracted attention in the areas of domain adaptation and cross-lingual transfer learning.
\citet{xquad} proposed to replace word embeddings in the PTLMs trained in the source or target languages.
\citet{inexpensive_da} proposed a vocabulary expansion using Word2Vec~\cite{word2vec} trained in the target domain for domain adaptation on a CPU. However, our preliminary experiments showed that simple replacement or vocabulary expansion harms performance in our setting because of the limited amount of data.
Unlike the previous studies, the proposed method requires no additional corpus by incorporating regularization of the word embeddings in the additional pre-training framework with the training data of the downstream task.

\section{Proposed Method}
The proposed method consists of three stages.

\paragraph{Additional Training of fastText}
First, we train a fastText model using the training data of the downstream task, 
where the model is initialized with publicly available fastText embeddings\footnote{\url{https://fasttext.cc/}}.

Our method introduces the embeddings of the PTLM vocabulary inferred by using the fastText model $F\in \mathbb{R}^{V_{\mathrm{LM}} \times d_{\mathrm{FT}}}$ as the word embeddings in the target domain. 
Unlike other word embedding methods such as GloVe~\cite{glove}, fastText retains subword information. Therefore, we can obtain the embeddings 
of the PTLM vocabulary containing subword units\footnote{We cannot obtain the embeddings of subwords shorter than the minimum length configured in fastText. 
The proposed method ignores such subwords in the additional training of the PTLMs.}.
The additional training of fastText runs much faster than the additional training of the PTLMs.
Note that TAPTER does not make any changes to the original vocabulary of the PTLMs.

\paragraph{Additional Pre-training of PTLMs}
Second, we train the entire PTLM using the training data of the downstream task. 
We input a token sequence $X \in V_{\mathrm{LM}}^l$ to the model.
We train the model with the loss function of language modeling $L_{\mathrm{LM}}(X)$ with $l_2$-norm regularization on the difference between the word embeddings.
That is, the loss function is 
\begingroup\makeatletter\def\f@size{10}\check@mathfonts
\begin{equation}
L_{\mathrm{LM}}(X) + \frac{1}{|R(X)|}\sum_{x_i \in R(X)} \|f(E_{x_i}) - F_{x_i} \|_2^2,
\label{eq:loss}
\end{equation}
\endgroup
where $R(X)$ is the set of the target tokens of the regularization.
The target tokens $R(X)$ exclude stop words and subwords shorter than the minimum length configured in fastText. 

The function $f$ maps a $d_{\mathrm{LM}}$-dimensional embedding to a $d_{\mathrm{FT}}$-dimensional embedding:
\[
f(z) =\mathrm{LN} (W_f z + b_f).
\]
LN denotes layer normalization \cite{layer_normalization}. $W_f \in \mathbb{R}^{d_{\mathrm{FT}} \times d_{\mathrm{LM}}}, b_f \in \mathbb{R}^{d_{\mathrm{FT}}}$ are trainable parameters.
The loss function of Eq.~\eqref{eq:loss} is designed to alleviate the catastrophic forgetting problem with the first term and to adapt the word embeddings to the target domain with the second term. \citet{l2_da1, l2_da2} proposed $l_2$-norm regularization for domain adaptation, but they did not incorporate it in the additional pre-training framework.

\paragraph{Fine-Tuning}
Finally, we run the standard fine-tuning process \cite{bert} without any regularization. 

\section{Evaluation}
\subsection{Dataset}
We evaluated the proposed method on two question answering datasets.
Table~\ref{tab:dataset} shows the statistics. The experimental setup is shown in Appendix~\ref{append:setting}.

\begin{table}[t]
\begin{center}
    \small
    \begin{tabular}{crrr}\hline
		        & Training  & Evaluation & Corpus Size \\ \hline
		        SQuAD1.1 & 87,599 & 10,570 & 2.62M \\
		        BioASQ5 & 4,950 & 150 & 1.38M \\ \hline
				\end{tabular}
	\caption{Statistics of the datasets. Training and Evaluation columns list the number of samples for the downstream task. Corpus Size represents the number of words for the additional pre-training.}
	\label{tab:dataset}
\end{center}
\end{table}

\paragraph{SQuAD} SQuAD1.1 is a task to answer a question with information from a textual source \cite{squad}.
The dataset provides pairs of a question and a related passage from Wikipedia as the textual source.
The input of the model is a token sequence that is a concatenation of the question and the passage.
The ground-truth answer is a span in the passage. 
The output of the model consists of the indices of the answer start and end positions.
The indices are calculated from the two-dimensional linear layer on top of the PTLM.
The official evaluation metrics are exact matching (EM) and partial matching (F1).

\paragraph{BioASQ} 
BioASQ5 is a question answering dataset in the biomedical domain \cite{bioasq}.
Following \citet{biobert}, we used the factoid questions pre-processed into SQuAD format.
We used three official evaluation metrics. 
Strict accuracy (SACC) is the rate at which the top-1 prediction is correct.
Lenient accuracy (LACC) is the rate at which the top-5 predictions include the ground-truth answer.
Mean reciprocal rank (MRR) is the average of the reciprocal of the rank of the ground-truth answer.
We trained the models with ten random seeds and report the average performance.
In the fine-tuning stage, as in~\citet{bioasq_train}, we first trained the model with SQuAD and then trained it with BioASQ.

\subsection{Compared Models}
We used three PTLMs, BERT-base-cased \cite{bert}, BioBERT \cite{biobert}, and RoBERTa-base \cite{roberta}.
BERT-base-cased was pre-trained with English Wikipedia (2.5B words) and BookCorpus (800M words) \cite{bookcorpus}.
BioBERT was initialized with BERT-base-cased and pre-trained with PubMed abstracts and PMC articles (18B words).
RoBERTa-base was pre-trained with 160GB corpora including news and Web articles as well as Wikipedia and BookCorpus (used to train BERT).
We compared three fine-tuning methods: standard fine-tuning, TAPT, and TAPTER.

\subsection{Results and Discussion}

\paragraph{Is TAPTER effective at adaptation to the biomedical domain?}
Table~\ref{tab:bioasq} shows the results in BioASQ.
We evaluated the performance of the domain adaptation 
with BERT-base-cased because the model does not use a biomedical corpus in the original pre-training.

TAPTER improved BERT's performance by 3.05/0.27/2.01 points (SACC/LACC/MRR) over the simple fine-tuning.
As well, TAPTER statistically significantly outperformed TAPT in the SACC (top-1 accuracy) and MRR metrics. 
We consider that the regularization of the word embeddings improves the adaptation of the PTLM.

Appendix~\ref{append:t-sne} shows the word embeddings of the models with principal component analysis.
The scatter plots show that the word embeddings of BERT-base-cased and TAPT resemble each other, though TAPTER and BioBERT have dissimilar word embeddings distributions to that of BERT-base-cased.
This indicates that the additional pre-training of language modeling alone does not adapt the static word embeddings to the biomedical domain unlike TAPTER.

\begin{table}[t]
\begin{center}
		\begin{tabular}{llll}\hline
		        & SACC & LACC & MRR  \\ \hline
		        BERT-base-cased & 37.88 & 54.00 & 43.87 \\
		        +TAPT & 39.47 & \textbf{54.27} & 44.69 \\
		        +TAPTER & \textbf{40.93**} & \textbf{54.27} & \textbf{45.88**} \\ \hline
		        BioBERT & 43.53 & \textbf{59.67} & 49.81 \\ 
		        +TAPT & \textbf{45.67} & 57.87 & \textbf{50.46} \\
		        +TAPTER & 44.60 & 58.33 & 50.02 \\ \hline
				\end{tabular}
	\caption{Performance on the development set of BioASQ5. Shown are the results of a paired $t$-test on the ten runs between TAPTER and TAPT (** : $p<.01$).}
	\label{tab:bioasq}
\end{center}
\end{table}

\paragraph{Is additional pre-training effective with the model pre-trained in the target domain?}
The additional pre-training from BioBERT did not improve the overall performance, although some of the scores slightly increased. 
There was no significant difference between TAPTER and TAPT in each metric ($p<.05$).
We consider that BioBERT has already learned the knowledge in the biomedical domain because it was pre-trained with a massive biomedical text.
Therefore, the additional pre-training had little effect on performance. 

\paragraph{Is TAPTER effective in the general domain?}
\begin{table}[t]
\begin{center}
		\begin{tabular}{lcc}\hline
		        & EM & F1  \\ \hline
		        BERT-base-cased & \textbf{79.12} & \textbf{87.55} \\
		        +TAPT & 78.42 & 87.12 \\
		        +TAPTER & 78.68 & 87.19 \\ \hline
		        RoBERTa-base & 82.76 & 90.40 \\ 
		        +TAPT & 83.01 & 90.45 \\
		        +TAPTER & \textbf{83.55} & \textbf{90.86} \\ \hline
				\end{tabular}
	\caption{Performance on the development set of SQuAD1.1.}
	\label{tab:squad}
\end{center}
\end{table}
Table~\ref{tab:squad} shows the results for SQuAD.
In the experiments with BERT neither of the additional pre-training methods improved performance.
On the other hand, in the experiments with RoBERTa, TAPTER improved performance by 0.79/0.46 points (EM/F1).
This was the best performance among the compared models on SQuAD.

We consider that TAPTER and TAPT improve performance when the corpora of the original pre-training were not dominated by in-domain data. A large part of the pre-training corpora of BERT is Wikipedia. Therefore, the additional pre-training was not effective. However, the pre-training corpora of RoBERTa cover broader topics. Although the corpora include Wikipedia, the additional pre-training can adapt the model to the Wikipedia domain. 

It is known that the performance of PTLMs tends to improve as the amount of pre-training corpora increases \cite{more_data, albert}. 
Our results show that TAPTER can improve the performance of PTLMs that were pre-trained with very large corpora even if the domain of the downstream task is included in the pre-training corpora. 

\paragraph{How well does TAPTER learn the language modeling and the word embeddings?}
\begin{figure}[t]
	\begin{center}
    \includegraphics[width=7.5cm]{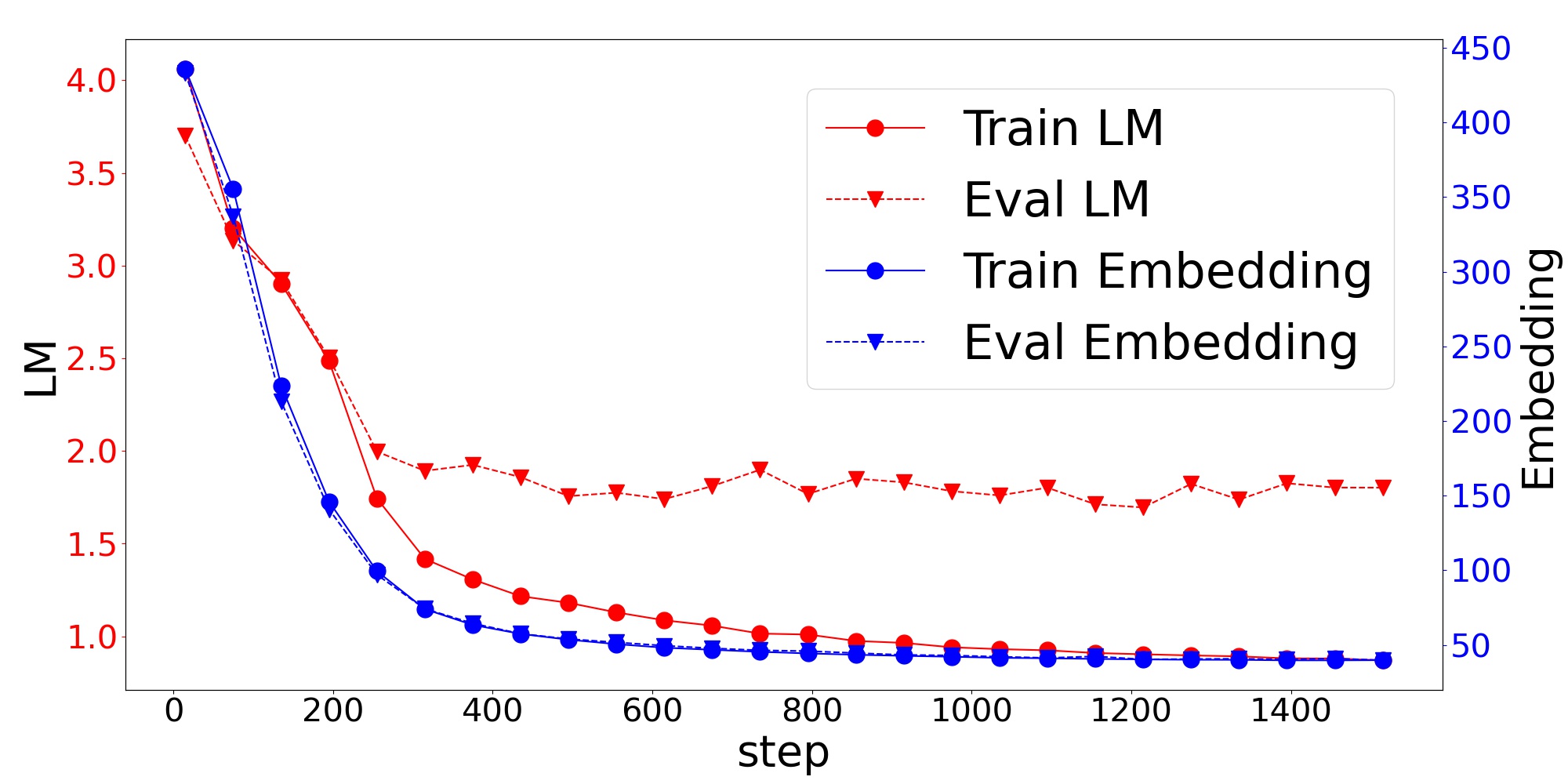}
	\caption{Learning curve. The left axis shows the first term in the loss. The right side shows the second term.}
	\label{fig:curve}
	\end{center}	
\end{figure}

Figure~\ref{fig:curve} shows the learning curve of the additional pre-training in BioASQ from BERT. We can see that the second term in Eq.~\eqref{eq:loss} representing the word embeddings decreased more sharply than the first term in Eq.~\eqref{eq:loss} representing the language modeling.
Since the BERT model is huge and complicated, we must learn the model slowly with a small learning rate over a large number of steps. However, the regularization term decreases quickly without corrupting the model. This is one of the advantages of TAPTER in low-resource settings.

In addition, the decrease in the first term on the development data stopped on the way.
However, the word embeddings were trained with less discrepancy between the training and development data.
We consider that the training data of BioASQ is too small to represent the distribution of the text in the biomedical domain. 
Since MLM takes a document-level sequence X as input, the search space of the true distribution $\Pr(X)$ is huge, and MLM is a very difficult task to train with limited training data. On the other hand, the regularization term depends on the word-level distribution $\Pr(x_i)$. Therefore, the model can decrease the regularization term on the evaluation data even in low-resource settings without overfitting.

\section{Conclusion}
We proposed a new fine-tuning process including additional pre-training with word embedding regularization.
TAPTER learns the meanings of words in the target domain by making the static word embeddings of the PTLM close to the word embeddings obtained in the target domain with fastText. 
TAPTER improves the performance of BERT in the biomedical domain. Moreover, 
it improves the performance of RoBERTa even in the Wikipedia domain although the original pre-training corpora of RoBERTa contain Wikipedia.

Many PTLMs with more parameters and trained with more data have been published \cite{t5,megatron}.
We believe that TAPTER is an important method to teach such largely pre-trained language models knowledge in the target domain. 

\bibliographystyle{acl_natbib}
\bibliography{theme}

\appendix
\section{Experimental Setup}
\label{append:setting}
We trained the models on one NVIDIA GeForce GTX 1080Ti (11GB). 
The hyperparameter settings are in Table \ref{tab:hyper1} and Table \ref{tab:hyper2}. 
The optimization algorithm was Adam~\cite{adam}.
We used PyTorch~\cite{pytorch} and Transformers~\cite{transformers}.
Stop words were implemented by NLTK~\cite{nltk}.
The word-level tokenizer was spaCy~\cite{spacy}.
For the target tokens of the regularization $R(X)$, we randomly selected 50 \% tokens in the input excluding stop words and subwords shorter than the minimum length configured in fastText. 
Following the default setting, the minimum length of the subwords in fastText was set to three. The maximum length was six.
In BioASQ, we lowercased the corpora in the additional training of fastText
and $R(X)$ in the additional pre-training because only a limited number of words containing capital characters appear.

Note that the computational time of our additional pre-training was about seven hours on one NVIDIA GeForce GTX 1080Ti (11GB) GPU, while that of BioBERT was more than ten days on eight NVIDIA V100 (32GB) GPUs.

\begin{table}[ht!]
\begin{center}
    \small
		\begin{tabular}{ccc}\hline
                & Pre-Training & Fine-Tuning \\ \hline
				 batch size & 256 & 128 \\
				 epochs & 100 & 5 / 2 / 10 \\
				 max seq. length & 512 & 384 \\
				 max query length & -- & 64 \\
				 learning rate & \multicolumn{2}{c}{5e-5} \\
				 warmup proportion & \multicolumn{2}{c}{0.06} \\
				 weight decay & \multicolumn{2}{c}{0.01} \\ \hline
				\end{tabular}
	\caption{Hyperparameters for the PTLMs. The numbers separated by slashes represent  SQuAD / the first stage of BioASQ / the second stage of BioASQ.}
	\label{tab:hyper1}\end{center}
\end{table}

\begin{table}[ht!]
\begin{center}
\small
	\begin{tabular}{ccc}\hline
		        & SQuAD & BioASQ \\ \hline
				 min count & 5 & 2 \\
				 epochs & 5 & 10 \\
				 dim & \multicolumn{2}{c}{300} \\ \hline
				\end{tabular}
	\caption{Hyperparameters for fastText. We used the default values for the hyperparameters not listed.}
	\label{tab:hyper2}
\end{center}
\end{table}

\section{Visualization of Word Embeddings}
\label{append:t-sne}

\begin{figure*}[t]
 \begin{minipage}{0.5\hsize}
 \begin{center}
    \includegraphics[clip, width=7.5cm]{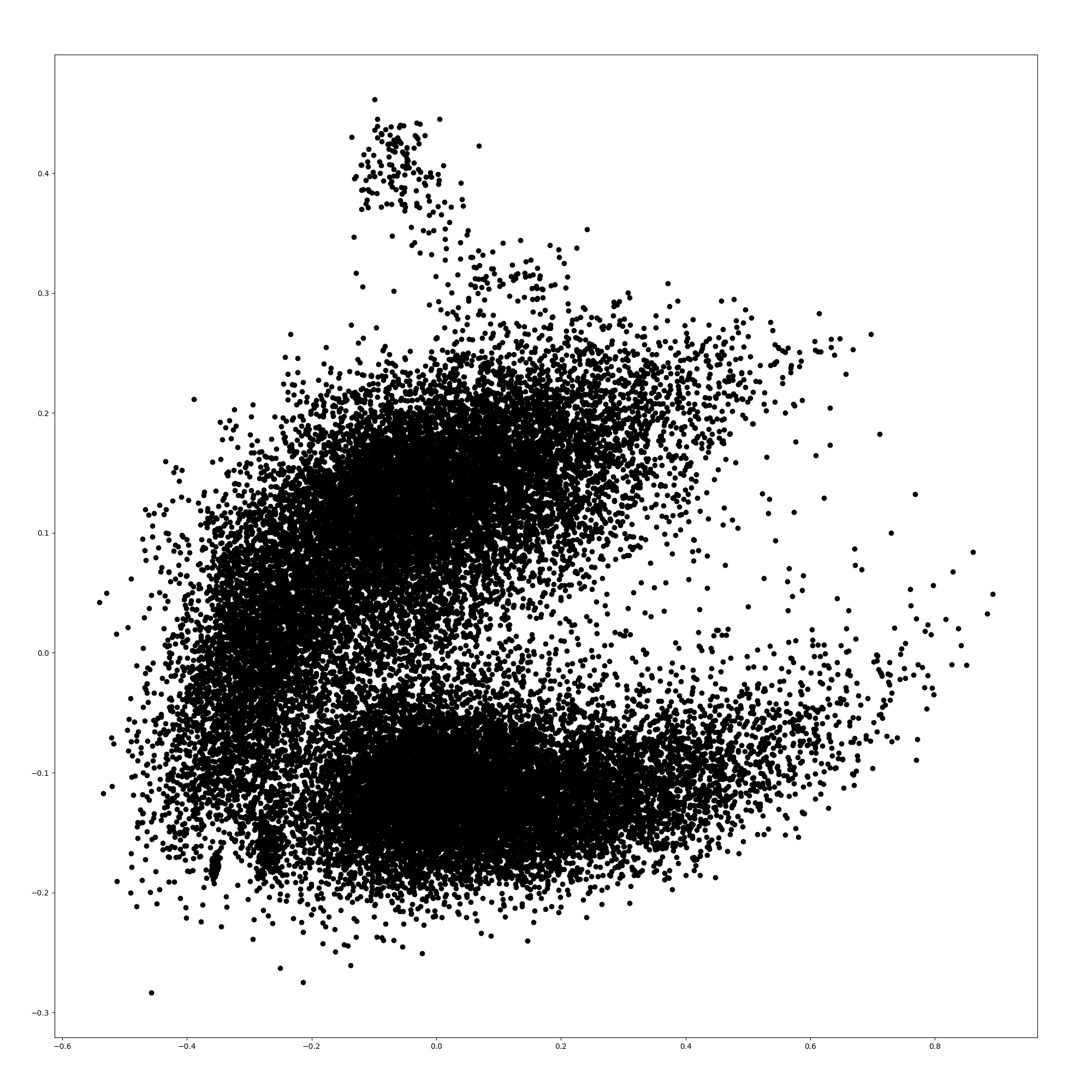}\\
	\caption{Word embeddings of BERT-base-cased}
	\label{fig:bert}
	\end{center}	
 \end{minipage}
 \begin{minipage}{0.5\hsize}
	\begin{center}
    \includegraphics[clip, width=7.5cm]{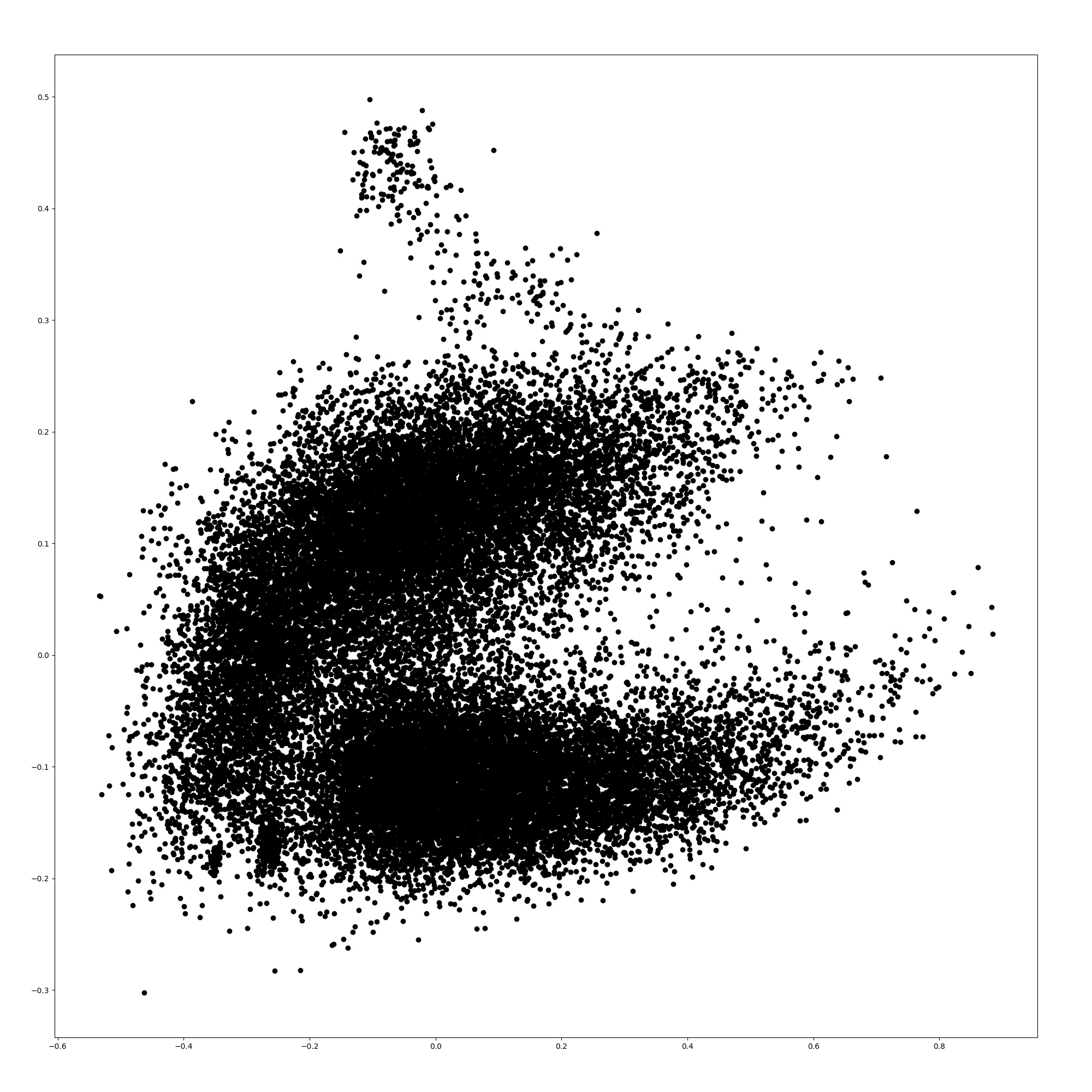}\\
	\caption{Word embeddings of model additionally pre-trained with TAPT}
	\label{fig:tapt}
	\end{center}	
 \end{minipage}
\end{figure*}

\begin{figure*}[t]
 \begin{minipage}{0.5\hsize}
 \begin{center}
    \includegraphics[clip, width=7.5cm]{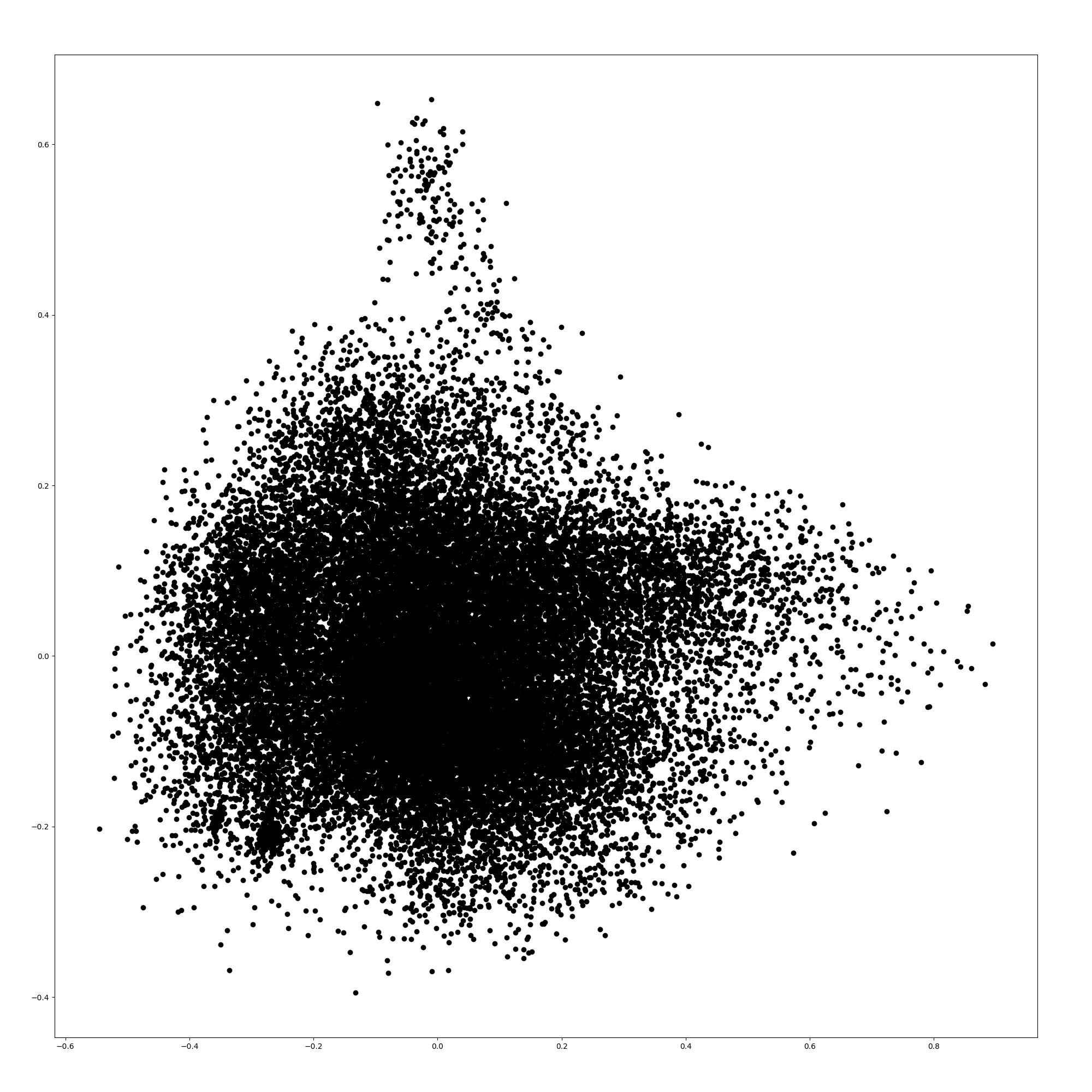}\\
	\caption{Word embeddings of model additionally pre-trained with TAPTER}
	\label{fig:tapter}
	\end{center}	
 \end{minipage}
 \begin{minipage}{0.5\hsize}
	\begin{center}
    \includegraphics[clip, width=7.5cm]{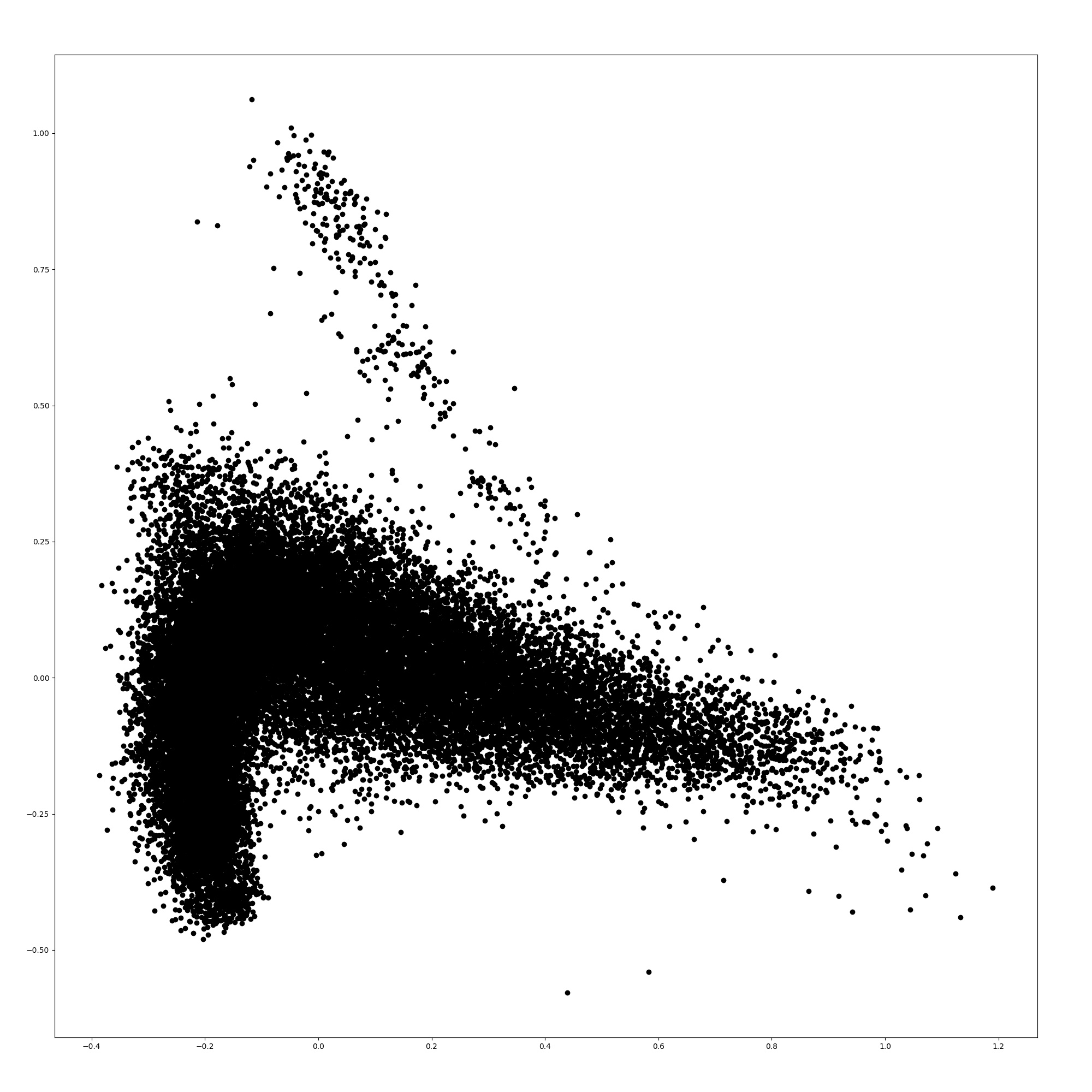}\\
	\caption{Word embeddings of BioBERT}
	\label{fig:biobert}
	\end{center}	
 \end{minipage}
\end{figure*}

Here, we show the word embeddings of the models with principal component analysis.
Figures~\ref{fig:bert}, \ref{fig:tapt}, \ref{fig:tapter}, and \ref{fig:biobert} are scatter plots of 
the word embeddings of BERT-base-cased, the model additionally pre-trained with TAPT, the model additionally pre-trained with TAPER, and BioBERT.

The figures show that the word embeddings of BERT-base-cased and TAPT resemble each other. The average distance between the embeddings of BERT-base-cased and TAPT among all words is 0.0576, although the distance between the embeddings of BERT-base-cased and TAPTER is 0.172. Therefore, the additional pre-training of language modeling alone does not adapt the static word embeddings to the biomedical domain unlike TAPTER.
TAPTER and BioBERT have dissimilar word embedding distributions to that of BERT-base-cased.

\clearpage


\end{document}